\documentclass[runningheads]{llncs}
\usepackage{graphicx}
\usepackage{tabularx}
\usepackage{amssymb}
\usepackage{hyperref}
\usepackage{arydshln} 
\usepackage{lipsum}
\usepackage{mdframed} 
\usepackage{xcolor}
\usepackage{soul}
\usepackage[normalem]{ulem}
\useunder{\uline}{\ul}{}

\begin{document}
\title{Employing Label Models on ChatGPT Answers Improves Legal Text Entailment Performance}
\titlerunning{Employing Label Models on ChatGPT Answers Improves Legal Text Entailment Performance}

\author{Chau Nguyen \and
Le-Minh Nguyen}
\authorrunning{C. Nguyen et al.}
%
\institute{Japan Advanced Institute of Science and Technology, Ishikawa, Japan
\email{chau.nguyen,nguyenml@jaist.ac.jp}}

\maketitle              
\begin{abstract}

The objective of legal text entailment is to ascertain whether the assertions in a legal query logically follow from the information provided in one or multiple legal articles.
ChatGPT, a large language model, is robust in many natural language processing tasks, including legal text entailment: when we set the \textit{temperature} = 0 (the ChatGPT answers are deterministic) and prompt the model, it achieves 70.64\% accuracy on COLIEE 2022 dataset, which outperforms the previous SOTA of 67.89\%.
On the other hand, if the \textit{temperature} is larger than zero, ChatGPT answers are not deterministic, leading to inconsistent answers and fluctuating results.
We propose to leverage label models (a fundamental component of weak supervision techniques) to integrate the provisional answers by ChatGPT into consolidated labels. By that way, we treat ChatGPT provisional answers as noisy predictions which can be consolidated by label models. 
The experimental results demonstrate that this approach can attain an accuracy of 76.15\%, marking a significant improvement of 8.26\% over the prior state-of-the-art benchmark.
Additionally, we perform an analysis of the instances where ChatGPT produces incorrect answers, then we classify the errors, offering insights that could guide potential enhancements for future research endeavors.

\end{abstract}

\section{Introduction}

Legal text entailment is a task in natural language processing (NLP) that involves determining whether a given statement logically follows from the facts stated in a legal text.
The development of automated systems for addressing the legal text entailment task is of critical significance, as it has the potential to provide substantial benefits to individuals with varying legal needs.
For example, it can help lawyers and legal professionals save time and effort analyzing large volumes of legal texts. Traditionally, lawyers have had to manually read and analyze legal documents to determine the relevant facts and legal arguments. With the help of automated legal text entailment systems, lawyers can quickly identify the most relevant information and arguments, which can help them make more informed decisions.
Besides, it is crucial for the development of advanced legal applications such as legal chatbots or legal question-answering systems. These applications can help make legal services more accessible and affordable, particularly for people who cannot afford expensive legal advice.

ChatGPT\footnote{https://chat.openai.com/} is a large language model developed by OpenAI\footnote{https://openai.com/} that is capable of understanding natural language text and generating human-like responses to prompts. Trained on a massive corpus of text data, ChatGPT has shown impressive performance across a wide range of natural language processing tasks, including language translation, summarization, and question-answering. We are interested in using ChatGPT to analyze legal texts, given its ability to understand the complex and nuanced language used in legal documents. Legal text entailment is one such task where ChatGPT's natural language processing capabilities can be particularly useful.

In the realm of weak supervision \cite{zhang2022survey}, a label model is a pivotal concept that serves as a crucial bridge between data with noisy labels and accurate model predictions. In the case where no gold data is available, and we only have the noisy labels from a variety of information sources like heuristics, expert rules, or crowdsourced annotations, the role of a label model is to generate probabilistic labels for the data by integrating these weak signals. 

In the context of using ChatGPT or similar language models, \textit{temperature} refers to a parameter that controls the randomness of the generated text. When generating text, higher \textit{temperature} values (e.g., 0.8 or 1.0) make the output more creative and varied, as the model is more likely to select less probable words and phrases. Lower \textit{temperature} values (e.g., 0.1 or 0.3) make the output more deterministic and focused, as the model tends to choose more probable words, resulting in more predictable responses. It is widely recognized that when the \textit{temperature} variable is set to a value other than zero, ChatGPT may yield inconsistent responses for the same prompt. In simpler terms, despite ChatGPT's strong language comprehension capabilities, it remains somewhat unpredictable and prone to variability. Consequently, we can regard the responses from ChatGPT as uncertain provisional answers. Hence, we propose to employ label models to refine these provisional answers and generate the final consolidated answers. Our experimental results demonstrate a significant improvement of 5.51\% compared to no legal model employed and 8.26\% compared to the previous state-of-the-art benchmark, implying that label models are suitable for integrating responses generated by large language models, such as ChatGPT.

In this paper, we preliminarily conduct experiments using the prompt-based configuration of ChatGPT to tackle the task of legal text entailment. The goal is to identify the most effective prompt type among three options: (i) \textit{Answer-only}, (ii) \textit{Answer-then-Explain}, and (iii) \textit{Reason-then-Answer}. The findings reveal that the \textit{Reason-then-Answer} prompt type outperforms the others. Specifically, ChatGPT with the \textit{Reason-then-Answer} prompt achieves a performance boost of 2.75\% on the COLIEE 2022 dataset \cite{kim2023coliee}, achieving an accuracy of 70.64\% compared to the previous accuracy of 67.89\%.

Subsequently, we employ ChatGPT (utilizing the \textit{Reason-then-Answer} prompt type) to generate multiple answers for each query, resulting in uncertain provisional predictions. To enhance these less-certain provisional predictions, we propose to employ label models to refine the results. The proposed strategy leads to a refined prediction that elevates the accuracy to 76.15\%, showcasing a substantial improvement of 8.26\% over the previous state-of-the-art benchmark.

Furthermore, we conduct an analysis of cases in which ChatGPT generates inaccurate responses. Subsequently, we categorize these errors, providing valuable insights that could inform potential improvements for future research endeavors.

\section{Related work}

\subsection{The COLIEE competition}

The development of automated systems for addressing legal text entailment is an emerging area of research that has the potential to revolutionize legal services. However, this field is still in its infancy, and much work remains to be done to develop accurate and efficient systems. To this end, the Conference on Legal Information Extraction and Entailment (COLIEE \cite{kim2023coliee}) has emerged as a prominent forum for advancing the development of automated legal text entailment systems. This annual international competition provides a platform for researchers and practitioners to showcase their latest advances in this field while promoting collaboration and knowledge sharing among participants.

\subsection{Approaches to legal text entailment in COLIEE competition}

The Conference on Legal Information Extraction and Entailment (COLIEE) has facilitated the development of a diverse range of approaches for the task of legal textual entailment. In COLIEE 2020 \cite{rabelo2021coliee}, participants employed a range of NLP techniques and models such as BERT \cite{devlin2018bert}, RoBERTa \cite{liu2019roberta}, GloVe \cite{pennington2014glove}, and LSTM \cite{hochreiter1997long}. The winning team, JNLP \cite{nguyen2020jnlp}, fine-tuned BERT-based models with Japanese legal data and utilized TF-IDF to achieve superior performance. Rule-based ensembles, SVM \cite{cortes1995support}, and attention mechanisms with word embeddings were also used to tackle the legal text classification task. In COLIEE 2021 \cite{rabelo2022overview}, the winning team HUKB \cite {yoshioka2021hukb} employed an ensemble of BERT models and utilized data augmentation, which outperformed the other approaches \cite{nguyen2021paralaw,wehnert2021legal,kim2021bm25,fujita2021predicate}. The 2022 competition saw further innovations, such as a method for selecting relevant parts from articles and employed an ensemble of BERT with data augmentation \cite{yoshioka2023hukb}, an ensemble of rule-based and BERT-based methods with data augmentation and person name inference \cite{fujita2023legal}, used the longest uncommon subsequence similarity comparison model \cite{lin2022rethink}, or employed an ensemble of graph neural networks with textbook nodes and sentence embeddings \cite{wehnert2023using}. These advances demonstrate the ongoing efforts to improve the performance of automated systems for legal text entailment, with significant implications for the future of legal services.

\subsection{Label models}
A label model, a fundamental component of weak supervision \cite{zhang2022survey} techniques, serves as a crucial tool in addressing the challenges posed by limited or noisy labeled data. In scenarios where obtaining accurate annotations is expensive or impractical, a label model offers an effective means of generating pseudo-labels from various sources of noisy supervision, such as heuristics, crowdsourcing, or distant supervision. Its primary purpose is to infer the true underlying labels of data points by leveraging the consensus or patterns present in the noisy annotations.

The necessity for label models arises from the increasing demand for robust and scalable methods in machine learning, especially when conventional manual labeling becomes prohibitively expensive or time-consuming. Label models bridge the gap by providing a systematic approach to harnessing the collective wisdom of multiple noisy sources, yielding more reliable labeled data for training models.

Label models operate by combining the input from different sources, applying statistical techniques to estimate the true labels probabilistically. Examples of label models include FlyingSquid model \cite{flyingsquid_fu2020fast}, Dawid-Skene model \cite{dawidskene_dawid1979maximum}, Hyper label model \cite{hyperlabelmodel_wu2022learning}, FABLE model \cite{fable_zhang2023leveraging}, and Generative model \cite{generativemodel_ratner2016data}.

\section{Methods}

\subsection{Preliminary experiment: Prompting ChatGPT for legal textual entaiment}

\begin{figure}[h]
  \centering
  \includegraphics[width=1.0\textwidth]{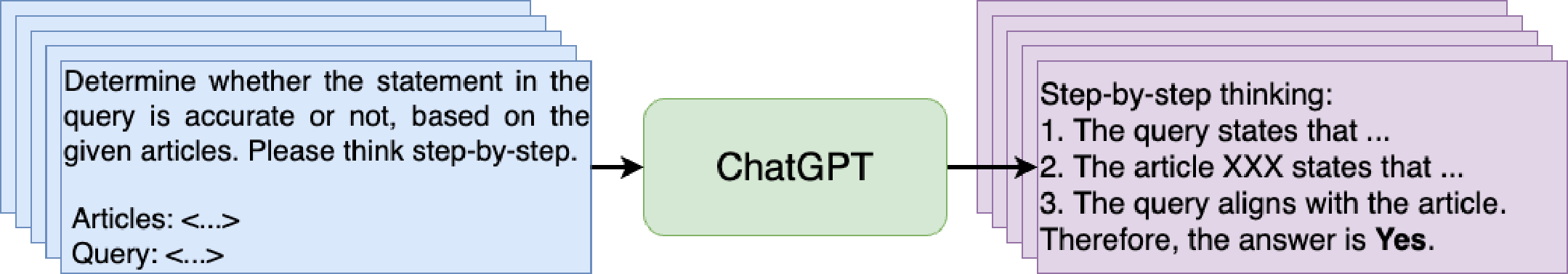}
  \caption{ChatGPT prompting procedure}
  \label{fig:chatgpt_overview}
\end{figure}

The utilization of Chain-of-Thought prompting \cite{chainofthought_wei2022chain} has the potential to encourage a more profound level of reasoning within a large language model, thereby leading to improved responses. However, the applicability of Chain-of-Thought prompting might not be suitable for all scenarios \cite{chen2023you}. In certain cases, asking ChatGPT to provide only the answer, without detailing each step of reasoning, could yield better outcomes. As a preliminary experiment, we would like to test ChatGPT with different types of prompts:
(i) \textit{Answer-only}, (ii) \textit{Answer-then-Explain}, and (iii) \textit{Reason-then-Answer} (similar to Chain-of-Thought prompting).
Figure \ref{fig:chatgpt_overview} provides an overview of the procedure for prompting ChatGPT to obtain an answer.

The designed prompts are as follows:

\begin{enumerate}
\item \textit{Answer-only}: ChatGPT only outputs the answer.

\begin{verbatim}
{
  "role": "user",
  "content": "Given a query (which is delimited with triple 
  backticks) and the related articles (which is also delimited 
  with triple backticks). Is the query entailed by the related 
  articles? Please provide a simple answer of either "Yes" or 
  "No", without any explanation.

  Query: ```{query}```

  Related articles: ```{related_articles}```"
}
\end{verbatim}

\item \textit{Answer-then-Explain}: ChatGPT outputs the answer and provides an explanation for its reasoning.

\begin{verbatim}
{
  "role": "user",
  "content": "Given a query (which is delimited with triple 
  backticks) and the related articles (which is also delimited 
  with triple backticks). Is the query entailed by the related 
  articles? Please provide the answer of "Yes" or "No", then 
  provide an explanation.

  Query: ```{query}```

  Related articles: ```{related_articles}```"
}
\end{verbatim}

\item \textit{Reason-then-Answer}: ChatGPT provides a step-by-step reasoning process and concludes with an answer.

\begin{verbatim}
{
  "role": "user",
  "content": "Given a query (which is delimited with triple 
  backticks) and the related articles (which is also delimited 
  with triple backticks). Is the query entailed by the related 
  articles? To answer, please use the following format:
    Step-by-step reasoning: <your step-by-step reasoning>
    Answer: <a clear "Yes" or "No" response>

  Query: ```{query}```

  Related articles: ```{related_articles}```"
}
\end{verbatim}

\end{enumerate}

When prompting ChatGPT, the \textit{temperature} parameter determines the level of randomness in the generated output. In this experiment, in order to ensure deterministic responses, we set the \textit{temperature} parameter to a value of 0.

We experiment with the test data of COLIEE 2022 \cite{kim2023coliee} and compare them with the previous systems' highest performances. For each test sample, the task involves assessing whether a given statement in a query can be inferred from the related legal articles provided in a list. The obtained experimental results are presented in Table \ref{tab:results}.

\begin{table}[h]
\centering
\caption{Results when prompting ChatGPT compared to previous methods}
\label{tab:results}
\begin{tabular}{|l|c|}
\hline
Method & Accuracy \\
\hline
\textit{Previous methods} & \\
~~~~KIS & 67.89\% \\
~~~~HUKB & 66.97\% \\
~~~~LLNTU & 60.55\% \\
~~~~OvGU & 57.80\% \\
~~~~UA & 54.13\% \\
~~~~JNLP & 53.21\% \\
\hline
\textit{Prompt type} & \\
~~~~\textit{Answer-only} & 66.97\% \\
~~~~\textit{Answer-then-Explain} & 67.89\% \\
~~~~\textit{Reason-then-Answer} & \textbf{70.64\%} \\
\hline
\end{tabular}
\end{table}

The results table demonstrates ChatGPT's robustness, with competitive accuracy compared to previous methods. In particular, ChatGPT using the \textit{Reason-then-Answer} prompt demonstrates a performance improvement of 2.75\%, reaching an accuracy of 70.64\% on the COLIEE 2022 dataset \cite{kim2023coliee}. This is in contrast to the previous accuracy of 67.89\%. Notably, the \textit{Reason-then-Answer} prompt yields the highest accuracy, indicating its appropriateness as a prompt type. 
One possible explanation for this outcome could be as follows: the \textit{Answer-only} prompt relies solely on the model's likelihood to predict tokens such as "Yes" or "No." In contrast, the \textit{Answer-then-Explain} approach, while furnishing explanations for the model's predictions, presents the answers upfront, with the subsequent explanation serving to support the anticipated response. Conversely, the \textit{Reason-then-Answer} approach offers a systematic, step-by-step analysis before arriving at a conclusion, closely emulating human reasoning processes. This emulation appears to contribute to the model's favorable performance.
Hence, we use the \textit{Reason-then-Answer} prompt for further experiments.

\subsection{Label models for integrating provisional ChatGPT answers into consolidated answers}
In these experiments, we set the \textit{temperature} variable to be non-zero to enable possibilities of ChatGPT producing different answers. As mentioned above, we use the \textit{Reason-then-Answer} prompt for prompting ChatGPT. We experiment with many values of \textit{temperature}: 0.1, 0.2, 0.3, ..., 0.9, 1.0. For each \textit{temperature}, we prompt ChatGPT 10 times. In other words, for each value of \textit{temperature}, there are 10 provisional answers for the queries in the dataset.

Table \ref{tab:stats} shows some information on the results of prompting ChatGPT 10 times with different \textit{temperature} values. It can be seen that within a single run, the best accuracy could be 76.15\% when the \textit{temperature} equals 0.4, 0.6, 0.8, 0.9 or 1.0. However, the corresponding \textit{min} accuracy could be down to 65.14\%, 66.06\%, or 67.89\%. It suggests that the best accuracy above may just because the model got lucky. We can look at the median values to demonstrate this point: the median values are 71.56\%, 71.10\%, 68.81\%, 72.02\%, 70.64\% for the \textit{temperature} equals 0.4, 0.6, 0.8, 0.9, 1.0, respectively. Another interesting observation is the differences of the performances of the runs (see the row \textbf{max-min}): while with \textit{temperature} equals 0.1, the difference is only 4.59\%, these values when the \textit{temperature} equals 0.2 to 1.0 ranges from 8.26\% to 12.84\%.

\begin{table}[h]
\centering
\caption{Accuracies when prompting ChatGPT 10 times with different \textit{temperature} values. Values in \textbf{bold}/\underline{underline} indicate the \textbf{highest}/\underline{lowest} value in each row.}
\label{tab:stats}
\begin{tabular}{|c|c|c|c|c|c|c|c|c|c|c|}
\hline
\textbf{Temperature} & \textbf{0.1}   & \textbf{0.2} & \textbf{0.3}   & \textbf{0.4}   & \textbf{0.5} & \textbf{0.6}   & \textbf{0.7} & \textbf{0.8}   & \textbf{0.9}   & \textbf{1.0}   \\ \hline
\textbf{max}         & 73.39          & 75.23        & 75.23          & \textbf{76.15} & {\ul 72.48}  & \textbf{76.15} & 75.23        & \textbf{76.15} & \textbf{76.15} & \textbf{76.15} \\ \hline
\textbf{min}         & \textbf{68.81} & 66.97        & {\ul 62.39}    & 65.14          & 64.22        & 65.14          & 65.14        & 66.06          & 67.89          & 66.06          \\ \hline
\textbf{max-min}   & {\ul 4.59}     & 8.26         & \textbf{12.84} & 11.01          & 8.26         & 11.01          & 10.09        & 10.09          & 8.26           & 10.09          \\ \hline
\textbf{avg}         & 71.83          & 71.47        & 69.45          & 71.74          & {\ul 69.72}  & 71.01          & 70.46        & 70.09          & \textbf{71.93} & 71.01          \\ \hline
\textbf{median}      & \textbf{72.48} & 71.56        & {\ul 68.81}    & 71.56          & 69.72        & 71.10          & 70.18        & {\ul 68.81}    & 72.02          & 70.64          \\ \hline
\end{tabular}
\end{table}

\begin{figure}[h]
  \centering
  \includegraphics[width=1.0\textwidth]{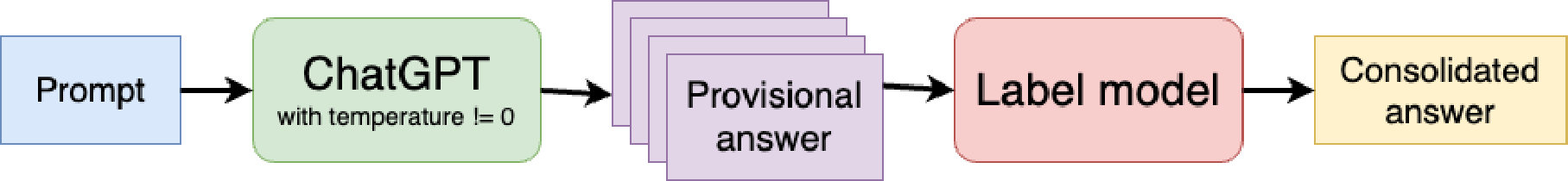}
  \caption{Employ label models on provisional answers produced by ChatGPT}
  \label{fig:main_fig}
\end{figure}

Based on those observations, it can be said that ChatGPT could achieve a high but fluctuating performance. Hence, we propose to treat ChatGPT answers as provisional answers, and to leverage the label models to integrate the provisional answers to consolidated answers (Figure \ref{fig:main_fig}).
We perform experiments for each of \textit{temperature} values above with different label models.
In particular, we experiment with the following label models:

\begin{itemize}
    \item Majority voting:  This is a simple label model where the predicted label for an example is determined by taking the majority vote from multiple provisional answers. It assumes that the majority decision is more likely to be correct.
    \item FlyingSquid \cite{flyingsquid_fu2020fast}: It employs both agreements and disagreements among the provisional answers to develop a labeling model that assesses the accuracy of the labeling functions. The answer with the highest probability is regarded as the final consolidated answer.
    \item Dawid-Skene \cite{dawidskene_dawid1979maximum}: The Dawid-Skene model is a probabilistic label model used in crowdsourcing scenarios. It estimates the true labels and the worker reliabilities in a crowdsourcing setting with multiple noisy annotators. In our case, the noisy annotations are the provisional answers produced by ChatGPT.
    \item Hyper label model \cite{hyperlabelmodel_wu2022learning}: The hyper label model is an analytical method for label integration that is constructed using a Graph Neural Network to ensure that its predictions remain unchanged or appropriately adjusted when the order of the provisional answers are changed.
    \item FABLE \cite{fable_zhang2023leveraging}: FABLE is a statistical label model which is built on a mixture of Bayesian label models. Each Bayesian label model here corresponds to a global pattern of correlation. The coefficients of the mixture components are predicted by a Gaussian Process classifier based on instance features.
    \item Generative model \cite{generativemodel_ratner2016data}: In weak supervision, a generative model can be used to estimate true labels through a process of “denoising” the provided provisional answers.
\end{itemize}

\begin{table}[h]
\centering
\caption{Results when employing label models on provisional answers. \textbf{LModel} means label model. \textbf{AVG} means the average value of a row/column. Values in \textbf{bold} indicate the \textbf{highest} value accuracy of each label model over all \textit{temperature} values. Values in \underline{\textbf{bold+underline}} indicate the highest average accuracy over all label models/all \textit{temperature} values.}
\label{tab:main_results}
\begin{tabular}{|l|c|c|c|c|c|c|c|c|c|c|l|c|}
\hline
\textbf{LModel\textbackslash Temp} & \textbf{0.1} & \textbf{0.2} & \textbf{0.3} & \textbf{0.4} & \textbf{0.5}   & \textbf{0.6} & \textbf{0.7} & \textbf{0.8} & \textbf{0.9} & \textbf{1.0} &  & \textbf{AVG}      \\ \hline
Majority voting                          & 72.48        & 73.39        & 70.64        & 73.39        & 74.31          & 73.39        & 74.31        & 73.39        & 73.39        & 71.56        &  & 73.03                 \\ \hline
FlyingSquid                              & 73.39        & 74.31        & 72.48        & 74.31        & \textbf{76.15} & 72.48        & 74.31        & 71.56        & 72.48        & 74.31        &  & 73.58                 \\ \hline
Dawid-Skene                              & 73.39        & 74.31        & 70.64        & 73.39        & 74.31          & 73.39        & 72.48        & 73.39        & 74.31        & 71.56        &  & 73.12                 \\ \hline
Hyper label model                        & 73.39        & 74.31        & 71.56        & 74.31        & \textbf{76.15} & 71.56        & 74.31        & 72.48        & 71.56        & 74.31        &  & 73.39                 \\ \hline
FABLE                                    & 72.48        & 73.39        & 68.81        & 73.39        & 71.56          & 73.39        & 72.48        & 73.39        & 74.31        & 70.64        &  & 72.38                 \\ \hline
Generative model                         & 74.31        & 73.39        & 72.48        & 73.39        & \textbf{76.15} & 73.39        & 74.31        & 71.56        & 74.31        & 75.23        &  & { \ul \textbf{73.85} }                \\ \hline
\textbf{\ \ \ \ \ \ \ \ AVG}                         & 73.24        & 73.85        & 71.10        & 73.70        & { \ul \textbf{74.77}} & 72.93        & 73.70        & 72.63        & 73.39        & 72.94        &  & \multicolumn{1}{l|}{} \\ \hline
\end{tabular}
\end{table}

Table \ref{tab:main_results} shows the results when employing label models on provisional answers. It can be seen that with \textit{temperature} = 0.5, the model could achieve an accuracy of 76.15\% quite consistently, with different label models: FlyingSquid, Hyper label model, and Generative model. For this case of \textit{temperature} = 0.5, the average accuracy over all label model is also the best when it achieves 74.77\%. The results suggest that employing label models is an appropriate method to integrate ChatGPT uncertain provisional answers. Besides, the results also suggest that the \textit{temperature} = 0.5 may be a good trade-off between the "creativity" value and the deterministic value for ChatGPT in our context. Moreover, the \textbf{AVG} column shows that the highest average accuracy over all label models is 73.85\% of the "Generative model" label model.

In summary, we found that the "Generative model" label model employed on 10 provisional ChatGPT answers with the \textit{temperature} = 0.5 achieves 76.15\% accuracy on the COLIEE 2022 legal text entailment dataset \cite{kim2023coliee}. This accuracy improves 5.51\% compared to a single ChatGPT prompting of 70.64\% and improves 8.26\% over the prior state-of-the-art benchmark of 67.89\%, suggesting the effectiveness of label models in integrating provisional answers.

We further investigate the performance of "Generative model" label model when integrating different numbers of provisional answers. In particular, instead of integrating 10 provisional answers (which achieves 76.15\%), we set the number of provisional answers to be from 3 to 9, and employ the "Generative model" label model, then report the average accuracy. For example, in case of three provisional answers provided, we consider all combinations of 3 answers out of 10 answers, employ the label model for each combination, then calculate the average accuracy. The results are shown in Table \ref{tab:ensemble_n_runs}. Interestingly, when the number of provisional answers equals 3, the performance is the lowest while the performance is highest when there are 8 provisional answers. We can also see a trend that, increasing the number of provisional answers, the performance tends to be higher and more stable.

\begin{table}[h]
\centering
\caption{Performance of "Generative model" label model when integrating different numbers of provisional answers. Values in \textbf{bold}/\underline{underline} indicate the \textbf{highest}/\underline{lowest} value in each row.}
\label{tab:ensemble_n_runs}
\begin{tabular}{|c|c|c|c|c|c|c|c|}
\hline
\textbf{\# provisional answers} & \textbf{3}  & \textbf{4} & \textbf{5}     & \textbf{6}     & \textbf{7} & \textbf{8}     & \textbf{9}     \\ \hline
\textbf{max}                    & {\ul 76.15} & 77.06      & \textbf{77.98} & \textbf{77.98} & 77.06      & \textbf{77.98} & \textbf{77.98} \\ \hline
\textbf{min}                    & {\ul 53.21} & 66.97      & {\ul 53.21}    & 67.89          & 68.81      & \textbf{70.64} & \textbf{70.64} \\ \hline
\textbf{avg}                    & {\ul 71.51} & 72.64      & 72.53          & 73.59          & 73.49      & \textbf{74.29} & 73.39          \\ \hline
\textbf{median}                 & {\ul 71.56} & 72.48      & 72.48          & 73.39          & 73.39      & \textbf{74.31} & 72.94          \\ \hline
\textbf{max-min}                & 22.94       & 10.09      & \textbf{24.77} & 10.09          & 8.25       & {\ul 7.34}     & {\ul 7.34}     \\ \hline
\end{tabular}
\end{table}

\section{Error analysis and classification}

To gain a better understanding of the errors made by ChatGPT for the \textit{Reason-then-Answer} prompt, we conducted an analysis over the mis-classified query by ChatGPT and classify the errors into four categories. These errors in ChatGPT's reasoning can be classified as follows:

\begin{itemize}
\item Category 1: Incorrect provision of facts by ChatGPT leading to incorrect conclusions. This category accounts for 15.62\% of the errors, indicating that ChatGPT has a problem with hallucinations, and it may make up facts to support its hallucinations. For instance, in the scenario below, ChatGPT fabricates a fact during the reasoning process in step 6 (highlighted in yellow), consequently arriving at an incorrect conclusion:

\begin{mdframed}[backgroundcolor=white, roundcorner=5pt]
    \textsf{\textcolor{red}{Related articles:}}
        
    \textsf{Article 143:}
    
    \textsf{(1) If a period is provided for in weeks, months, or years, the period is calculated in accordance to the calendar. }
    
    \textsf{(2) If a period does not commence at the beginning of the week, month, or year, that period expires on the day immediately preceding the day which corresponds to the commencement day in the last week, month or year; provided, however, that if the period is provided for in months or years and the last month does not contain a corresponding day, the period expires on the last day of that month.}
    
    \vspace{\baselineskip}
    
    \textsf{\textcolor{red}{Query:}}
    
    \textsf{If an agreement is made at 11:00 a.m. on July 12 of a given year that the obligation will be fulfilled within one month, the period expires at 12:00 p.m. on August 13 of the same year, both of which do not fall on Sundays, holidays as provided in the Act on National Holidays, or any other holidays.}
    
    \vspace{\baselineskip}
    
    \textsf{\textcolor{red}{ChatGPT step-by-step reasoning:}}
    
    \textsf{1. The query states that if an agreement is made on July 12 of a given year that the obligation will be fulfilled within one month, the period expires at 12:00 p.m. on August 13 of the same year.}
    
    \textsf{2. The query specifies that the period does not fall on Sundays, holidays as provided in the Act on National Holidays, or any other holidays.}
    
    \textsf{3. Article 143(1) states that if a period is provided for in months, the period is calculated in accordance with the calendar.}
    
    \textsf{4. Article 143(2) states that if a period does not commence at the beginning of the month, that period expires on the day immediately preceding the day which corresponds to the commencement day in the last month.}
    
    \textsf{5. Since the agreement is made on July 12 and the period is for one month, the period will end on August 12.}
    
    \textsf{\hl{6. Since August 13 is not a Sunday, holiday as provided in the Act on National Holidays, or any other holiday, the period can be extended to August 13.}}
    
    \textsf{7. Therefore, the query is entailed by the related articles.}
    
    \vspace{\baselineskip}
    
    \textsf{\textcolor{red}{ChatGPT answer:} Yes}
    
    \textsf{\textcolor{red}{Gold answer:} No}
    
\end{mdframed}

\item Category 2: The inability of ChatGPT to draw correct conclusions from correct premises. This category accounts for 65.62\% of the errors, demonstrating that ChatGPT still has limitations in understanding legal texts and reasoning legal scenarios. For instance, in the following scenario, ChatGPT can accurately identify the premises presented in both the query and the related article; however, it fails to draw an appropriate conclusion:

\begin{mdframed}[backgroundcolor=white, roundcorner=5pt]
    \textsf{\textcolor{red}{Related articles:}}
        
    \textsf{Article 548:}
    
    \textsf{The right to cancel is extinguished if the holder of the right to cancel, intentionally or negligently, causes significant damage to, or makes it impossible to return the object of the contract, or converts the object into another kind of thing by processing or alteration; provided, however, that this does not apply if the holder of the right to cancel does not know of the holder's right to cancel.}
    
    \vspace{\baselineskip}
    
    \textsf{\textcolor{red}{Query:}}
    
    \textsf{If the holder of the right to cancel negligently makes it impossible to return the object of the contract, the right to cancel is extinguished even if the holder of the right to cancel does not know of the holder's right to cancel.}
    
    \vspace{\baselineskip}
    
    \textsf{\textcolor{red}{ChatGPT step-by-step reasoning:}}
    
    \textsf{1. The related article states that the right to cancel is extinguished if the holder of the right to cancel intentionally or negligently makes it impossible to return the object of the contract.}
    
    \textsf{2. The related article also states that this provision does not apply if the holder of the right to cancel does not know of their right to cancel.}
    
    \textsf{3. The query states that if the holder of the right to cancel negligently makes it impossible to return the object of the contract, the right to cancel is extinguished even if the holder of the right to cancel does not know of their right to cancel.}
    
    \textsf{\hl{4. Therefore, the query is entailed by the related article.}}
    
    \vspace{\baselineskip}
    
    \textsf{\textcolor{red}{ChatGPT answer:} Yes}
    
    \textsf{\textcolor{red}{Gold answer:} No}
\end{mdframed}

\item Category 3: Difficulties faced by ChatGPT while reasoning on "mutatis mutandis" articles. This category accounts for 9.38\% of the errors, suggesting that ChatGPT has limitations in linking the non-obvious claims. In the following case, Article 558 pertains to contracts for sale, while Article 559 indicates that these provisions also apply to contracts for value other than contracts for sale with necessary modifications. However, ChatGPT fails to deduce the application of mutatis mutandis in this context:

\begin{mdframed}[backgroundcolor=white, roundcorner=5pt]
    \textsf{\textcolor{red}{Related articles:}}
        
    \textsf{Article 558:}
    
    \textsf{The expenses associated with contracts for sale are borne equally by both parties.}

    \textsf{Article 559:}
    
    \textsf{The provisions of this Section apply \hl{mutatis mutandis} to contracts for value other than contracts for sale; provided, however, that this does not apply when it is not permitted by the nature of the contract for value.}
    
    \vspace{\baselineskip}
    
    \textsf{\textcolor{red}{Query:}}
    
    \textsf{The expenses associated with conclusions of lease contracts are borne equally by both parties.}
    
    \vspace{\baselineskip}
    
    \textsf{\textcolor{red}{ChatGPT step-by-step reasoning:}}
    
    \textsf{1. The query is about the expenses associated with lease contracts.}
    
    \textsf{2. Article 559 states that the provisions of the section apply to contracts for value other than contracts for sale.}
    
    \textsf{3. However, it also states that this does not apply when it is not permitted by the nature of the contract for value.}
    
    \textsf{4. Article 558 states that the expenses associated with contracts for sale are borne equally by both parties.}
    
    \textsf{\hl{5. Since the query is specifically about lease contracts and there is no mention of lease contracts in the related articles, we cannot infer that the expenses associated with conclusions of lease contracts are borne equally by both parties. Therefore, the answer is No.}}
    
    \vspace{\baselineskip}
    
    \textsf{\textcolor{red}{ChatGPT answer:} No}
    
    \textsf{\textcolor{red}{Gold answer:} Yes}
\end{mdframed}

\item Category 4: The lack of relevant articles in the dataset prevents ChatGPT from reaching conclusive answers or may lead to incorrect responses. This category accounts for 9.38\% of the errors, suggesting that the data could be improved in the future. In the following case, the lack of sufficient related articles led to confusion in ChatGPT, resulting in the model's inability to reach a conclusion. In this case, the articles 140, 141, 142 should be added to the list of related articles.

\begin{mdframed}[backgroundcolor=white, roundcorner=5pt]
    \textsf{\textcolor{red}{Related articles:}}
        
    \textsf{Article 143:}
    
    \textsf{(1) If a period is provided for in weeks, months, or years, the period is calculated in accordance to the calendar.}

    \textsf{(2) If a period does not commence at the beginning of the week, month, or year, that period expires on the day immediately preceding the day which corresponds to the commencement day in the last week, month or year; provided, however, that if the period is provided for in months or years and the last month does not contain a corresponding day, the period expires on the last day of that month.}
    
    \vspace{\baselineskip}
    
    \textsf{\textcolor{red}{Query:}}
    
    \textsf{If an agreement is made at 5:00 p.m. on January 30 of a non-leap year that the obligation would be fulfilled within one month, the period expires at 12:00 p.m. on February 28 of the same year, both of which do not fall on Sundays, holidays as provided in the Act on National Holidays, or any other holidays.}
    
    \vspace{\baselineskip}
    
    \textsf{\textcolor{red}{ChatGPT step-by-step reasoning:}}
    
    \textsf{1. The query states that if an agreement is made on January 30 of a non-leap year that the obligation would be fulfilled within one month, the period expires at 12:00 p.m. on February 28 of the same year.}
    
    \textsf{2. Article 143(1) states that if a period is provided for in months, the period is calculated in accordance with the calendar.}
    
    \textsf{3. Article 143(2) states that if a period does not commence at the beginning of the month, the period expires on the last day of that month.}
    
    \textsf{4. Since the agreement in the query is made on January 30 and the period is for one month, the period would expire on February 28.}

    \textsf{\hl{5. The query specifies that the period would expire at 12:00 p.m. on February 28, but neither Article 143(1) nor Article 143(2) mention anything about a specific time of day.}}

    \textsf{6. However, the query also specifies that the expiration date must not fall on Sundays, holidays as provided in the Act on National Holidays, or any other holidays, which is not addressed in the related articles.}

    \textsf{\hl{7. Therefore, it cannot be definitively concluded whether the query is entailed by the related articles.}}
    
    \vspace{\baselineskip}
    
    \textsf{\textcolor{red}{ChatGPT answer:} No}
    
    \textsf{\textcolor{red}{Gold answer:} Yes}
\end{mdframed}

\end{itemize}

\section{Conclusion}

In summary, our proposed method involves employing label models to integrate the provisional answers produced by ChatGPT into consolidated answers. The results in our experiments show that employing the "Generative model" label model to 10 provisional ChatGPT answers, with a \textit{temperature} value of 0.5, yields an accuracy of 76.15\% in our task. This showcases a notable enhancement of 8.26\% compared to the previously established state-of-the-art benchmark. Furthermore, we conduct an analysis of situations in which ChatGPT provides inaccurate responses. Subsequently, we categorize these errors, providing valuable insights that could direct potential improvements for future research endeavors.

\bibliographystyle{splncs04}
\bibliography{references}

\end{document}